\documentclass[letterpaper, 10 pt, conference]{ieeeconf} % for review 
\IEEEoverridecommandlockouts
\IEEEoverridecommandlockouts                              % This command is only needed if 
\usepackage{times}
\usepackage{epsfig}
\usepackage{graphicx}
\usepackage{amsmath}
\usepackage{amssymb}
\usepackage{booktabs}
\usepackage{makecell}
\usepackage{adjustbox}
\usepackage[dvipsnames]{xcolor}
\usepackage{multirow}
\usepackage{hyperref}
\usepackage{multicol}
\usepackage[symbol]{footmisc}
\usepackage{cite}
\usepackage{textcomp}
\usepackage{soul}

\title{\LARGE \bf Center Direction Network for Grasping Point Localization on Cloths}

\author{Domen Tabernik$^{* 1}$, Jon Muhovi\v{c}$^{1}$, Matej Urbas$^{1}$ and Danijel Sko\v{c}aj$^{1}$% <-this % stops 
\thanks{$^{*}$Corresponding author.}%
\thanks{$^{1}$Members of the Faculty of Computer and Information Science, University of 
Ljubljana, Ve\v{c}na pot 113, Ljubljana, Slovenia {\tt\small domen.tabernik@fri.uni-lj.si}.}%
}

\begin{document}

\maketitle              
\thispagestyle{empty}   % REMOVE for final
\pagestyle{empty}       % REMOVE for final

%%%%%%%%% ABSTRACT
\begin{abstract}

Object grasping is a fundamental challenge in robotics and computer vision, critical for advancing robotic manipulation capabilities. Deformable objects, like fabrics and cloths, pose additional challenges due to their non-rigid nature. In this work, we introduce CeDiRNet-3DoF, a deep-learning model for grasp point detection, with a particular focus on cloth objects. CeDiRNet-3DoF employs center direction regression alongside a localization network, attaining first place in the perception task of ICRA 2023's Cloth Manipulation Challenge. Recognizing the lack of standardized benchmarks in the literature that hinder effective method comparison, we present the ViCoS {Towel} Dataset. This extensive benchmark dataset comprises 8,000 real and 12,000 synthetic images, serving as a robust resource for training and evaluating contemporary data-driven deep-learning approaches. Extensive evaluation revealed CeDiRNet-3DoF's robustness in real-world performance, outperforming state-of-the-art methods, including the latest transformer-based models. Our work bridges a crucial gap, offering a robust solution and benchmark for cloth grasping in computer vision and robotics. Code and dataset are available at: {\url{https://github.com/vicoslab/CeDiRNet-3DoF}}.%and \url{https://go.vicos.si/clothdataset}.}
\end{abstract}

% KEYWORDS only for FINAL VERSION
%\begin{IEEEkeywords}
%List of keywords (from the RA Letters keyword list)
%\end{IEEEkeywords}

%%%%%%%%% BODY TEXT
\section{Introduction}

In robotics and computer vision, object grasping is a fundamental challenge crucial for advancing manipulation capabilities. {The variability in shapes of real-world objects requires advanced perception algorithms for precise grasp point detection, particularly challenging with deformable materials like fabrics due to their non-rigid nature, leading to non-rigid transformations and self-occlusions.}

In addressing cloth manipulation, traditional robotics research often focuses on physical grasping or manipulation strategies (e.g., folding/unfolding) and simplifies the environment using conventional computer vision techniques~\cite{Hou2017}, still present in recent works~\cite{Caporali2020}. Recently, more robust deep-learning techniques have been employed to address perception. Some approaches manage it by directly regressing one or two grasp points using neural networks~\cite{Saxena2019,Corona2018,Seita2018}, while others treat grasp point estimation as a segmentation problem~\cite{Ren2021,Qian2020} for edges, hems, collars, or sleeves.
A significant limitation in current research is the lack of standardized benchmarking, as each study evaluates its solution on a private dataset, hindering comparison between approaches. Efforts to standardize benchmarking~\cite{Garcia_Camacho2020,Garcia_Camacho2022a}, such as the recent \emph{Cloth Manipulation and Perception Competition}~\cite{Garcia_Camacho2022} from ICRA, still lack a comprehensive dataset for training and evaluating modern data-driven deep-learning methods.

\begin{figure}
    \centering
    \includegraphics[width=0.95\linewidth]{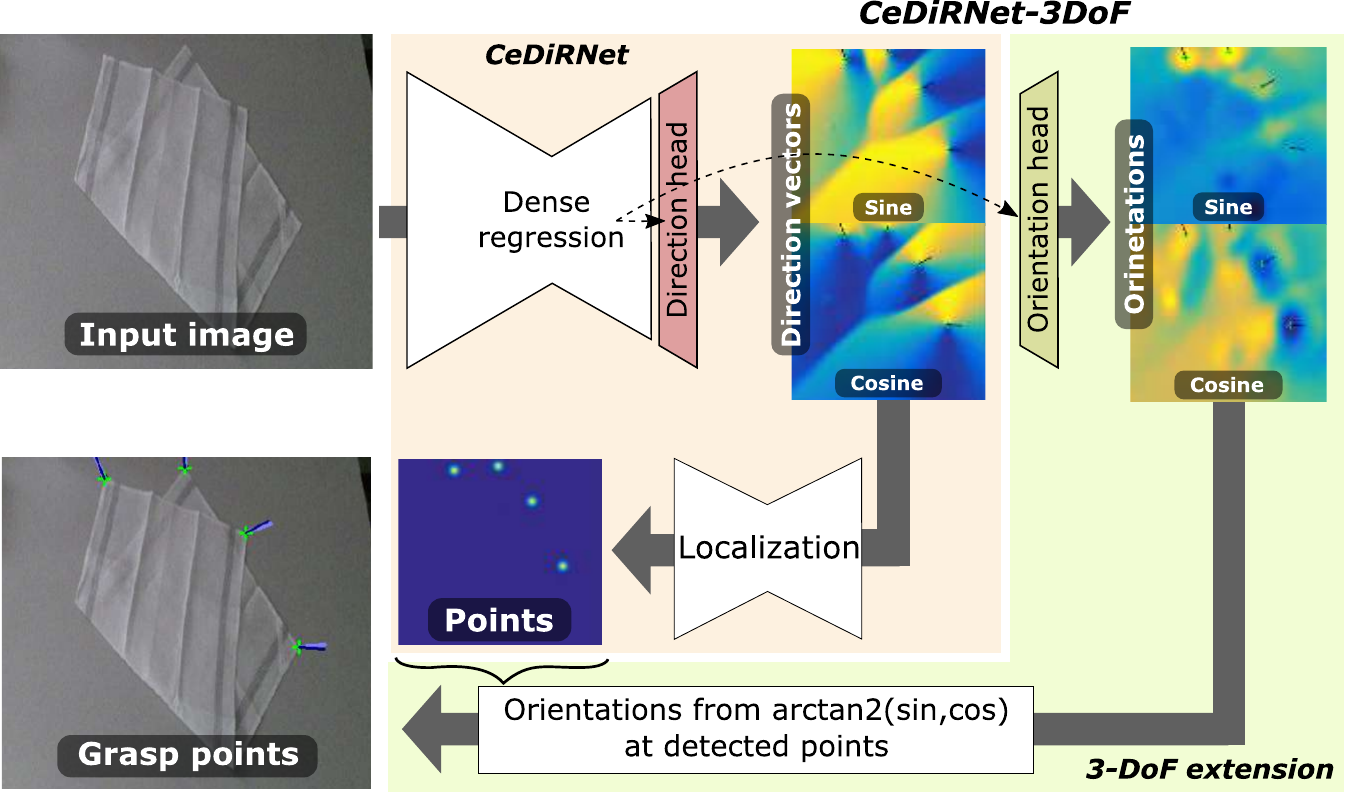}
    \caption{The proposed CeDiRNet-3DoF architecture.}
    \label{fig:architecture}
\end{figure}
 
While computer vision algorithms have demonstrated remarkable capabilities in addressing grasping challenges, the absence of a dedicated benchmark hinders effective method comparison. Our motivation stems from this critical need, driving us to propose a novel benchmark dataset and a novel method designed to detect grasp points in cloth objects. {In this paper, we focus on grasp points for towel manipulation tasks (folding/unfolding) following the definition of grasp points from~\cite{Garcia_Camacho2022}, where only visible physical corners are considered valid graspable points while grasp orientation is a single dimensional angle-of-approach defined relative to the image plane.}
Our contributions in this paper are twofold. First and foremost, we introduce a deep-learning methodology for identifying grasp points on deformable objects {by extending our previously introduced} CeDiRNet~\cite{Tabernik2024PR} { with 3-DoF extension}. The proposed CeDiRNet-3DoF, which achieved the first place in perception task at the \emph{2\textsuperscript{nd} Cloth Manipulation and Perception Competition} from ICRA 2023, enables accurate grasp point detection through regression of center-direction and a separate localization network {as part of the original CeDiRNet} (see, Fig.~\ref{fig:architecture}), while {the proposed 3-DoF extension enables estimation of optimal grasp orientation (i.e., angle-of-approach) and simultaneously improving grasp point detection.} 

To address the dearth of benchmark datasets in this domain, our second contribution is an extensive, publicly available dataset comprising 8,000 real images built {using ten diverse towels} from  Household Cloth Objects~\cite{Garcia_Camacho2022a} and 12,000 synthetic ones. The proposed dataset, termed the \emph{ViCoS {Towel} Dataset}, encompasses {images of diverse towel configurations, backgrounds, lighting conditions, and occluding object clutter captured on a tabletop from a viewpoint simulating the robotic perception}, thus serving as a benchmark for training and evaluating modern deep-learning methods in grasp point detection. The proposed dataset facilitates a rigorous evaluation of CeDiRNet-3DoF and its comparison with ten state-of-the-art deep-learning methods, including several recent transformer-based approaches. 

The remainder of this paper is structured as follows: Section~\ref{sec:related_work} discusses related work, and Section~\ref{sec:method} presents the methodology. The ViCoS {Towel} Dataset is introduced in Section~\ref{sec:dataset} and used for evaluation and benchmarking in Section~\ref{sec:experiments}. We conclude with a discussion in Section~\ref{sec:conclusion}.

\section{Related work}
\label{sec:related_work}

In the field of grasp point estimation for deformable objects, traditional approaches have historically relied on highly engineered features~\cite{Maitin-Shepard2010}. Grasp point estimation methods were employed in combinations with Bag of Features, sliding windows, SVM, and engineered measures to define ``grasp goodness''~\cite{Ramisa2012}. Others also used Difference-of-Gaussians and Gabor features coupled with Harris corners~\cite{Doumanoglou2016a}. Traditional approaches also persist in recent work for estimating grasp points from point cloud segmentation using wrinkledness measures~\cite{Caporali2020}.

While traditional approaches are known for not requiring extensive training data, they often lack robustness for real-world applications. Consequently, deep-learning methods have been increasingly explored. Yamazaki et al.~\cite{Yamazaki2018} applied deep-learning segmentation and VGG19 features for gripper position estimation in towel folding, albeit with a small dataset of 275 images. Demura et al.~\cite{Demura2018} employed pre-trained YOLO and transfer learning for grasp point estimation on folded towels with a dataset of 200 images. Seita et al.~\cite{Seita2018} also used YOLO with additional grasp point layers for the bed-making task, achieving superior performance compared to the Harris corner detector on a larger dataset of 2,000 images. Corona et al.~\cite{Corona2018} used a hierarchical tree of CNNs for cloth classification and estimating the 1st and 2nd grasp points. They used 2,000 real and 6,000 synthetic images generated from four garments; however, their dataset lacks public availability. Saxena et al.~\cite{Saxena2019} employed semantic segmentation of garments and specialized CNN for each garment for grasp point estimation. They proposed a large-scale dataset comprising 18k real and 95k synthetic images but did not make it publicly available.

More recent works continue to leverage deep-learning techniques for grasp point estimation. Ren et al.~\cite{Ren2021} learned a model without real data using synthetic depth data and domain adaptation, focusing on segmenting graspable regions. Zhu et al.~\cite{Zhu2023} proposed BiFCNet for semantic segmentation of graspable regions, training on NYU-Depth v2, and evaluating on a lab coat. Garcia-Camacho et al.~\cite{Garcia_Camacho2020,Garcia_Camacho2022a} recently proposed the Household Cloth Object Set for benchmarking deformable object manipulation but lacks suitable datasets for training and evaluation. Lips et al.~\cite{Lips2024} recently introduced a large-scale dataset for learning and evaluating modern deep-learning approaches. However, only 2,000 real images are proposed, leaving only synthetic images of sufficient quantity for training, albeit generated for high-level photo-realism with Blender.

A notable limitation in the literature is the absence of a comprehensive comparison among approaches, often coupled with specific robotic systems that hinder replication of results. Many methods are evaluated on proprietary datasets, limiting accessibility and scalability for contemporary deep-learning techniques. Addressing these challenges is an important objective as we strive to provide solutions for benchmarking and comparing deep-learning methodologies in the field. 

\section{Methods}\label{sec:method}

This section details our proposed method for grasp point localization termed CeDiRNet-3DoF that leverages our previously introduced CeDiRNet~\cite{Tabernik2024PR}. Method adapted specifically for the regression of graspable points and their corresponding angles-of-approach for robotic arms. We start by presenting the formulation of the original CeDiRNet as a reference {(previous work)}, then define our novel {3-DoF extension} designed for the regression of grasp points and angles.

\subsection{CeDiRNet}

The original Center Direction Network from~\cite{Tabernik2024PR} comprises of two network modules: i) a dense regression of center-directions and ii) a localization network. We offer a concise overview of each module and refer the reader to~\cite{Tabernik2024PR} for a more comprehensive description.

\paragraph{Dense regression of center-directions}
In the first module, the network regresses directions to the closest target point from all its surrounding pixel locations. A direction angle $\phi_{i,j} = \tan \big( \frac{m-j}{n-i} \big)$ is defined for each pixel location ${\mathbf{x}=(i,j)}$ pointing towards the closest target point ${\mathbf{y}=(n,m)}$. A direction angle $\phi$ is not regressed directly but instead predicted as a re-parameterized angle in {${C_{\sin}=sin(\phi)}$ and ${C_{\sin}=\cos(\phi)}$}, therefore regressing two fields:
{
\begin{align}
    C_{\sin},C_{\cos} &= DenseRegNet(\mathcal{I}),
\end{align}
}
where $\mathcal{I}\in\mathbb{R}^{N\times M}$ is the input image, {$DenseRegNet$ is a regression network with two dense output fields} and $C_{\sin},C_{\cos}\in\mathbb{R}^{N\times M}$ are two dense output fields corresponding to $\sin(\cdot)$ and $\cos(\cdot)$ of direction angle $\phi$ as depicted in top middle images in Fig.~\ref{fig:architecture}. 
As proposed in~\cite{Tabernik2024PR}, the original regression network uses an encoder-decoder model, with ResNet-101 for the encoder backbone and Feature Pyramid Network for the decoder{, where the last part of decoder contains a dense output head with convolutions and upsampling for the increased resolution output (Conv2D 3x3, GroupNorm, ReLu, Bilinear upsampling, Conv2D 3x3)}. 

The center-direction module is learned with an L1 loss function: 
\begin{equation}
\label{eq:polar_loss}
\begin{aligned}
    \mathcal{L}_\phi =& \sum_{i,j}W_\varepsilon(\mathbf{x}_{i,j}) \cdot |C_{\sin}(\mathbf{x}_{i,j}) - \hat{C}_{\sin}(\mathbf{x}_{i,j})| + \\ 
                 & \sum_{i,j}W_\varepsilon(\mathbf{x}_{i,j}) \cdot |C_{\cos}(\mathbf{x}_{i,j}) - \hat{C}_{\cos}(\mathbf{x}_{i,j})|,
\end{aligned}
\end{equation}
where the groundtruth data are two dense fields $\hat{C}_{\sin}$ and $\hat{C}_{\cos}$ constructed from $\sin(\cdot)$ and $\cos(\cdot)$ respectively for every pixel location, indicating the direction of the pixel towards the closest point to be detected.  The loss function has an additional per-pixel weight $W_\varepsilon(\mathbf{x}_{i,j})$ with a cutoff distance threshold $\varepsilon$ to balance the loss between different ground-truth points and between ground-truth points and the background to ensure pixels for every ground-truth point have equal importance in the learning process. 

\paragraph{Localization from center-directions}

The second module is used to extract the exact point locations from the regressed center-directions { fields}, which is implemented as an additional deep neural network ${\mathcal{O}_{cent} = {LocNet}(C_{\sin}, C_{\cos})}$, outputting probability map for grasp points. Finding local-maxima values in the resulting $\mathcal{O}_{cent}$ yields a list of locations of detected points.

${LocNet}$ is implemented as a lightweight hourglass architecture consisting of four levels in both encoder and decoder and a small number of channels (16 or 32). The localization network is domain agnostic {and can be trained on synthetic data since each problem domain is always effectively translated to a common appearance} of re-parameterized center direction angles that are the same regardless of the actual visual appearance of the objects. For more details, the reader is referred to~\cite{Tabernik2024PR}.

\subsection{CeDiRNet-3DoF for grasp point localization}

{We now detail the proposed 3-DoF extension for the CeDiRNet, resulting in the CeDiRNet-3DoF formulation that enables the application to the grasp-point localization problem}. Grasp points are defined with their location $\mathbf{y}$ in the image and their {single-dimensional} angle-of-approach $\theta$ for the robotic manipulator. {Finding grasp point location is implemented as regression of center-direction angles $\phi$ in sin/cos form pointing towards $\mathbf{y}$ using the original CeDiRNet architecture, while we now propose to reuse the same regression model to also} predict the angle-of-approach $\theta$ for each predicted grasp point. For every pixel location $\mathbf{x}_{i,j}=(i,j)$, we assign an angle-of-approach value $\theta_{i,j}$ based on its closest corresponding ground-truth grasp point. Similarly to center-direction, we do not directly regress angle-of-approach $\theta$ but instead re-parameterize it with {$D_{\sin}=\sin(\theta)$ and $D_{\cos}=\cos(\theta)$  to avoid regressing periodic angle values while also normalizing to $[-1,1]$. This follows the same re-parametrization from the original CeDiRNet for regressing center directions~\cite{Tabernik2024PR}.}
The regression network now predicts two additional fields $D$ related to $\theta$:

{
\begin{equation}
\begin{aligned}
    (C_{\sin},C_{\cos}), (D_{\sin}, D_{\cos}) &= DenseRegNet(\mathcal{I}),
\end{aligned}
\end{equation}
}
where $C_{\sin},C_{\cos}\in\mathbb{R}^{N\times M}$ are dense output fields for center-direction angles $\phi$, and $D_{\sin},D_{\cos}\in\mathbb{R}^{N\times M}$ are dense output fields for angle-of-approach $\theta$.

\paragraph{Training} {The regression network ${DenseRegNet}$ is trained to simultaneously predict all four dense output fields using the same network. However, we use a separate dense output head for angle-of-approach $\theta$ to allow for a certain degree of specialization since the output of center direction angle $\phi$ significantly differs from the output of angle-of-approach $\theta$. Therefore, all encoder and decoder weights are shared except for the final dense output heads.} The loss function $\mathcal{L}$ consists of a loss for center-direction $\mathcal{L}_\phi$ and a loss for angle-of-approach $\mathcal{L}_\theta$. We additionally balance both losses using uncertainty weighing~\cite{Kendall2018}, resulting in the final loss:
\begin{align}
    \mathcal{L} &= \frac{1}{2\sigma_{\phi}^2} \mathcal{L}_\phi + \frac{1}{2\sigma_{\theta}^2} \mathcal{L}_\theta + \log\sigma_{\phi}\sigma_{\theta},
\end{align}
where weighting values $\sigma_{\phi}$ and $\sigma_{\theta}$ are additional learnable parameters defining the uncertainty weight for each corresponding loss, thus eliminating the need for manually setting additional hyperparameter weights {and preventing the domination of gradients from one task to negatively influencing the other task.}

Both losses, $\mathcal{L}_\phi$ and $\mathcal{L}_\theta$, use L1 { following the Eq.~\ref{eq:polar_loss}, where $\mathcal{L}_\theta$ is applied to $(D_{\sin}, D_{\cos})$ fields instead of $(C_{\sin}, C_{\cos})$}; however, loss for angle-of-approach $\mathcal{L}_\theta$ is not computed in the background as is in $\mathcal{L}_\phi$. Instead, we compute $\mathcal{L}_\theta$ only within a cutoff distance $\varepsilon$ area around ground-truth grasp points ($30\times30$ pixels) and ignore loss for pixels further away from ground-truth grasp points {since they are not relevant and would only distract from learning relevant pixels around grasp points}. Note that we ignore the background only during training; during inference, we regress for all pixels {but use only ones at detected grasp points.}

We train only the regression network and use a pre-trained generic localization network from~\cite{Tabernik2024PR} {since the appearance of regressed center-direction angles in sin/cos form (which is input for localization net) does not change across different problem domains, therefore eliminating the need to fine-tune localization network for the specific domain, as proposed in~\cite{Tabernik2024PR}.}

\paragraph{Inference} During inference, we apply {generic} localization network to the regressed center-direction fields $(C_{\sin}, C_{\cos})$ to predict grasp point locations. For each predicted grasp point, we estimate its angle-of-approach by converting the regressed angle-of-approach value at the detected location from $\sin$/$\cos$ to actual angle $\hat{\theta} = \tan^{-1}\Big(\frac{D_{\cos}}{D_{\sin}}\Big)$. 

\paragraph{Comparison to the original CeDiRNet} 
Compared to the original CeDiRNet~\cite{Tabernik2024PR}, the overall architecture of CeDiRNet-3DoF remains the same with only two main networks: a) a regression network and b) a localization network. However, as part of our 3-DoF extension, we introduce several changes to the regression network: i) replace ResNet-101 encoder backbone with ConvNext, ii) add additional dense output head for regressing sin/cos of angle-of-approach $\theta$, and iii) add uncertainty weights~\cite{Kendall2018} for balancing center-direction $\mathcal{L}_\phi$ and angle-of-approach $\mathcal{L}_\theta$ losses.

\begin{figure*}
    \centering
    \includegraphics[width=1.0\linewidth]{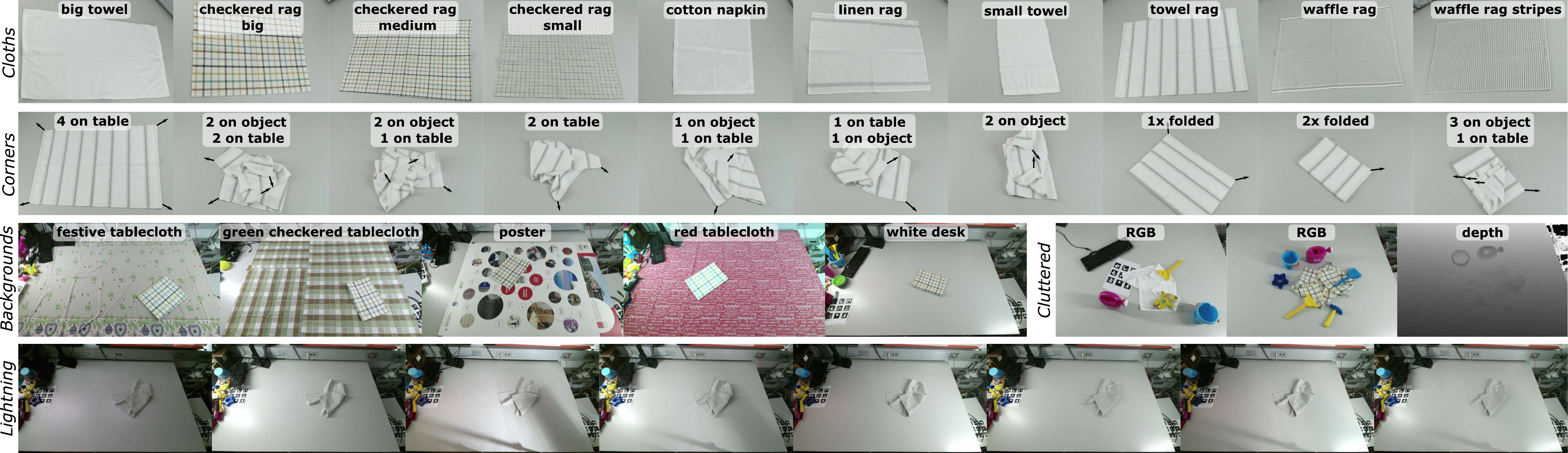}
    \caption{Overview of images in the ViCoS {Towel} Dataset with different towels, corner configurations, backgrounds, clutter, and lightning.}
    \label{fig:database_overview}
\end{figure*}

\section{The ViCoS {Towel} Dataset}\label{sec:dataset}

We propose a novel benchmark dataset for grasp point localization on cloth objects, termed the ViCoS {Towel} Dataset. The dataset consists of images depicting various towels on top of the desk from the perspective of a robotic view that was systematically captured in various conditions using Kinect V2. This resulted in 8,000 annotated high-resolution images ($1920\times1080$) with depth information (RGB-D), which is sufficient for both training deep-learning models and for extensive benchmarking under various conditions. Examples are depicted in Fig.~\ref{fig:database_overview}, including a depth in the third row.

\subsection{Dataset construction}

\begin{table}
    \centering
    \caption{Configurations (* objects not present in the training set).}
    \label{tab:configurations}
    \begin{tabular}{ll}
    \toprule
        \textit{\makecell[l]{Object/cloth\\types}} &          
        \makecell[l]{big towel; small towel; checkered rag big;\\
                     checkered rag medium; checkered rag small*;\\
                     cotton napkin*; linen rag; towel rag;\\
                     waffle rag; waffle rag stripes} \\
    \midrule                                                     
        \textit{\makecell[l]{Object\\positions}} &     
        \makecell[l]{4 on table; 2 on object 2 on table; 2 on object; \\
                     on table; 1 on table on 1 on object; 1 on object \\
                     1 on table; 3 on object 1 on table; 2 on object; \\
                     1x folded (2 visible); 2x folded (1 visible)} \\
    \midrule
        \textit{\makecell[l]{Backgrounds}} &
        \makecell[l]{festive tablecloth*; green checkered tablecloth;\\
                     white desk; poster; red tablecloth} \\
    \midrule        
        \textit{\makecell[l]{Lighting\\conditions}} & 
        \makecell[l]{no lights; left light; strong right light;\\
                     strong right light + left light; weak right light; weak \\
                     right light + left light; both right lights; all lights} \\
    \bottomrule
    \end{tabular}
\end{table}

We systematically varied various cloth objects (i.e., towels), configurations of visible grasp points (i.e., corners of the towels), backgrounds, lighting conditions, and the presence of clutter in the background in all possible combinations. A detailed description of each condition is provided below.

\paragraph{Objects} Proposed dataset contains objects from the Household Cloth Object Set~\cite{Garcia-Camacho2022}. In particular, we included 10 towels that were part of the \emph{2\textsuperscript{nd} Cloth Manipulation and Perception Challenge} at ICRA 2023 (see the first row in Table~\ref{tab:configurations} and Fig.~\ref{fig:database_overview}). We consider each { visible physical} corner of the towel as a potential graspable point that is the subject of detection and localization, according to the rules of the challenge.

\paragraph{Object position} Each object was captured in 10 different positions lying on the tabletop, which simulated different levels of visibility of individual grasp points. This included fully spread-out objects with all corners visible, crumpled objects with some corners hidden, and folded objects with overlapping corners. Note that \textit{1 on cloth, 1 on table} has a similar configuration to \textit{1 on table, 1 on cloth}, but the examples are more crumpled. All configurations are listed in the second row in Table~\ref{tab:configurations} and depicted in the second row in Fig.~\ref{fig:database_overview}.

\paragraph{Backgrounds} Five different tabletop cloths or objects were used to increase the variance of the background as listed in the third row in Table~\ref{tab:configurations} and shown in the third row in Fig.~\ref{fig:database_overview}.

\paragraph{Lighting conditions} Dataset images were captured under eight different lighting conditions. We used several lights positioned around the table (two lights on the right, a strong and a weak one, and one on the left), which were turned on and off in different combinations to obtain several different scenes with i) poor lights, ii) mild shadows, iii) strong shadows, and iv) fully illuminated scene without shadows. Light conditions are listed in the last row in Table~\ref{tab:configurations} and depicted in the last row in Fig.~\ref{fig:database_overview}. Images with different lighting conditions were captured automatically; therefore, the same scene with the exact same object position, background, and clutter was captured in eight different lighting settings.

\paragraph{Clutter} Finally, we captured images with and without additional clutter in the scene. Various clutter objects were added either on the desk and around the cloth object or directly on the cloth themselves to create additional difficult scenes with occlusion of the cloth (see right side of the third row in Fig.~\ref{fig:database_overview}).

\subsection{Annotations} 
All visible corners of the cloth object were manually annotated in all images. Corners were annotated with a point label and represented a potential grasping point that needs to be detected and localized. We also annotated the angle-of-approach for each corner at 45° outwards relative to the cloth side edges. This resulted in a total of 20,784 annotation points and angles.

\subsection{Training and testing splits} We held out two towels (\emph{checkered rag small} and \emph{cotton napkin}) and one background (\emph{festive tablecloth}) object from the training set, which are then used exclusively for the testing, thus ensuring fair evaluation with objects that were not present in training. All remaining configurations (corners, lighting positions, w/ and w/o clutter) are included in both the training and testing set. This resulted in 5,120 training and 2,880 testing images. 

\subsection{Synthetic training data}
We also provide 12,000 additional synthetic images as part of the proposed ViCoS {Towel} Dataset that can be used to increase the diversity of the training data. Synthetic images were generated using MuJoCo\footnote{\href{https://mujoco.org/}{https://mujoco.org/}} simulation environment, where each image depicts a towel positioned on the table {at 80~cm distance at 70-90° viewing angle} in random configurations based on physics simulation, using random image texture for both towel and backgrounds {(12 towel and 10 background textures). Each towel was resized to between 24 and 45~cm in each dimension and simulated with grid-positioned ellipsoid elements at~3 cm intervals. We determined its final position by simulating a drop from 70~cm above the ground, randomly applying forces to the edges and corners to crumple the cloth into a complex shape. For a diverse set of shadows, directional lighting was positioned 1 m above the ground with random directions and color temperatures between 2000° and 6500° K.}

\section{Experiments}\label{sec:experiments}

The evaluation was carried out on the ViCoS {Towel} Dataset, and the method was benchmarked against ten state-of-the-art deep-learning approaches. We also conducted an ablation study to substantiate {our design choices} and the advantages of incorporating synthetic data and depth information.

\begin{figure*}
    \centering
    \includegraphics[width=0.92\linewidth]{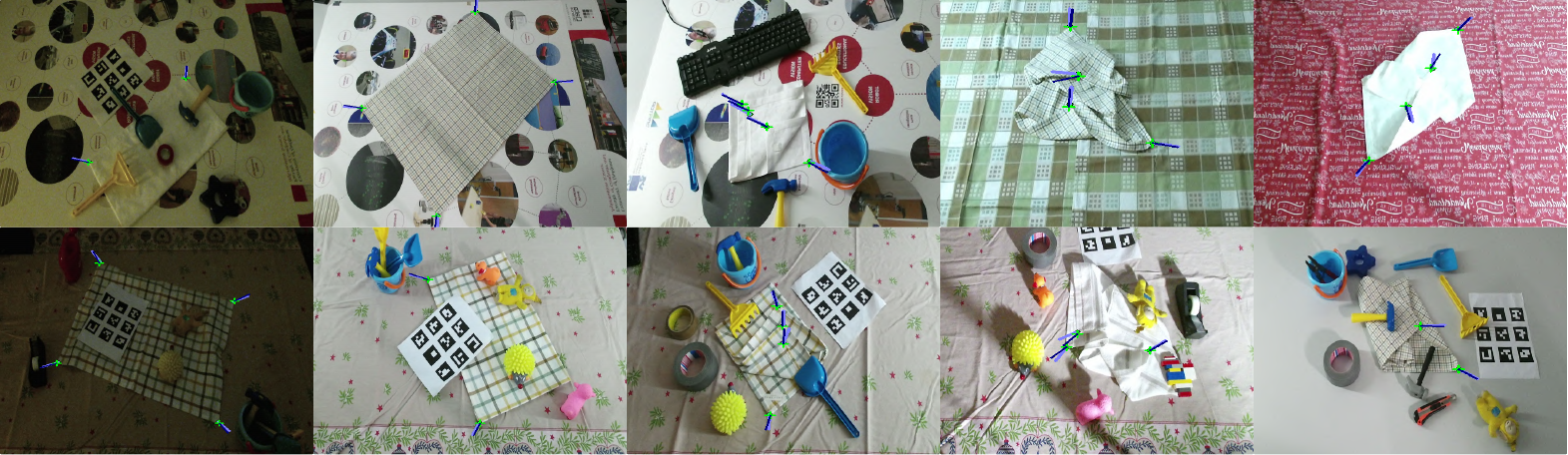}
    \caption{Several examples of correct grasp point detection (green cross) and estimation of angle-of-approach (dark lines predicted, light lines ground-truth).}
    \label{fig:detection_examples}
\end{figure*}

\subsection{Implementation details}

We trained separately on synthetic and real data, using weights from the synthetic training as initialization for real-data training. We trained with a single GPU, a batch size of 4, and a learning rate of $10^{-4}$ (Adam optimizer) for 10 epochs using a polynomial decay (power of 0.9). Fixed image sizes were used for training and testing: $512\times512$ for synthetic data and $768\times768$ for real data, { to approximately match the difference in object scales in different scenes.} Two data augmentation techniques were also applied—Gaussian blur ($\sigma=[0.5, 2]$) and color jittering (randomly selected in range $[-0.3, 0.3]$ for brightness, contrast, saturation, and hue), each with a probability of 0.5. For the backbone, we used the ConvNext-B model in all experiments unless explicitly stated otherwise. Where applicable, we also normalized the depth values to the range $[0,1]$. CeDiRNet-3DoF achieves an inference speed of approximately 20 FPS for half HD resolution on NVIDIA A100, and surpasses 8 FPS for full HD resolution.

\begin{table}
  \centering 
  \caption{CeDiRNet-3DoF on ViCoS {Towel} Dataset with ablation study. \label{tab:cedirnet_main}}
  \setlength{\tabcolsep}{2.5pt}
  \begin{tabular}{l|ccccccc} %ccc}
    \toprule
    \textit{\makecell[l]{Uncertainty weighing}} & & & & \checkmark  & \checkmark & \checkmark  &  \checkmark  \\
    \textit{\makecell[l]{3-DoF dense output head }} & & & \checkmark & \checkmark  & \checkmark & \checkmark  &   \checkmark \\
    \textit{\makecell[l]{Regressing angle-of-approach}} & & \checkmark & \checkmark & \checkmark  & \checkmark & \checkmark  & \checkmark \\
    \textit{\makecell[l]{With pre-trained synthetic data}} & \checkmark & \checkmark & \checkmark & \checkmark  & \checkmark &  &  \\
    \textit{RGB-D} & \checkmark & \checkmark & \checkmark & \checkmark  &  & \checkmark  & \\

    \midrule
    \midrule
    \textit{Precision [\%]} \textbf{\textuparrow} & 79.4 & 82.4 & \textbf{83.2} & 83.0 & 81.5 & 79.5 & 80.4 \\
    \textit{Recall [\%]} \textbf{\textuparrow} & 80.1 & 82.5 & 84.3 & \textbf{84.6} & 80.2  &  73.0 &  72.0 \\
    \textit{F1 [\%]} \textbf{\textuparrow}  & 77.0 & 80.0 & 81.2 & \textbf{81.4} & 78.0 & 73.0 & 72.7 \\
    \textit{\makecell[l]{Localization error [px]\textbf{\textdownarrow}}} & 1.7 & 1.7 & 1.7  &  \textbf{1.6}  & 1.7  & 1.9 & 1.7 \\
    \textit{\makecell[l]{Orientation error [°]\textbf{\textdownarrow}}} & - & 6.5 & 6.5 & \textbf{6.4} & 6.7 & 6.5  & 6.5 \\
    \bottomrule
  \end{tabular}  
\end{table}

\subsection{Evaluation metrics}

The evaluation focused on point localization performance in the context of grasping. We measured two sets of metrics: i) metrics for grasp point detection and ii) metrics for localization error and angle-of-approach error. A grasp point is correctly detected if the prediction is within 20 pixels of the ground truth, corresponding to 2-4 cm for objects in the dataset. If multiple predictions are near one ground truth, only one is counted as correct; the others are false positives. For correctly detected grasp points, we report localization error in pixels (translation error) and angle-of-approach error in degrees (rotation error).

\subsection{Overall results}
Table~\ref{tab:cedirnet_main} presents the results of CeDiRNet-3DoF's application to the ViCoS {Towel} Dataset for grasp point estimation. When using proposed 3-DoF extension with RGB-D data and pre-trained on synthetic data, our optimal variant achieves a precision of 83.0\%, recall of 84.6\%, and an F1 measure of 81.4\% in grasp point detection. With smaller distance tolerances, F1 reduces slightly to 80.4\% at 10 px (1-2 cm), and 76.1\% at 5 px (0.5-1 cm). The localization error is approximately 1.6 pixels, with an angle-of-approach error ranging between 6-7°. If exclusively testing on novel cloths and backgrounds (360 images), the performance mirrors the one achieved on the entire test set, maintaining an F1 measure of around 81.1\%. Several examples of good performance in difficult situations are depicted in Fig.~\ref{fig:detection_examples}.

\paragraph{Ablation study}
We also assessed several design choices (uncertainty weights, dense output head for 3-DoF, disabling angle-of-approach regression) and performance without synthetic data and depth information. Results in the left side of Table~\ref{tab:cedirnet_main} show our design choices' positive impact: uncertainty weights improve the F1 measure by 0.2 percentage points (pp), and the dense output head for 3-DoF adds 1.2 pp compared to using a shared head between center-directions and angle-of-approach. Not regressing the angle-of-approach reduces the F1 measure by 3 pp, suggesting that angle-of-approach regression positively influences center-direction regression, possibly acting as regularization and preventing overfitting. Combining all our design choices also slightly positively influences localization error (+0.1 pixels) and the orientation error (+0.1°).

The right side of Table~\ref{tab:cedirnet_main} shows the benefits of synthetic data and depth information. Excluding these leads to a significant performance drop. Without synthetic pre-training, the F1 measure decreases by 8-9 percentage points (pp), and the positive impact of depth information is most evident when synthetic data is used.

\subsection{Parametric performance study}

We conducted a supplementary parametric performance study on individual towels, corner configurations, backgrounds, and clutter. Results, sorted by F1 score, are reported in Table~\ref{tab:cedirnet_main}. CeDiRNet-3DoF performs best when the cloth is laid flat with all corners visible and worst with folded cloths, as folded edges appear as corners but should not be detected. Different towels and background types result in smaller variances compared to corner configurations. The highest performance is with \textit{towel rag} or \textit{checkered rag medium} towels and the \textit{festive tablecloth} background, while the \textit{checkered rag small} towel and \textit{red tablecloth} background perform worst. Interestingly, the best background was not in the training set, highlighting the absence of overfitting. Minimal overfitting is also apparent with the two towels not in the training set, attaining 77-80\% F1 score, which is close to 81.4\% in all test samples, indicating sufficient diversity of towels in the dataset to learn a generalized model. Clutter shows minimal impact, demonstrating the robustness to background clutter, which can occlude grasp points.

\begin{table}
  \centering  
  \caption{Parametric study for CeDiRNet-3DoF with RGB-D.\\(*~objects not present in the training set; Cltr=Clutter)
  \label{tab:parametric_study}}
  \setlength{\tabcolsep}{5pt}
  \begin{tabular}{m{0.1cm}lccccc}
    \toprule
    \hspace{-10pt}& \textit{} & \textit{\makecell[c]{Prec.\\~[\%]}} & \textit{\makecell[c]{Recall\\~[\%]}} & \textit{\makecell[c]{F1\\~[\%]}} & \textit{\makecell[c]{Loc.\\~[px]}} & \textit{\makecell[c]{Orient.\\~[°]}} \\
    \midrule
    \multirow{10}{*}{\rotatebox[origin=c]{90}{Corner configuration}} & 4 on table & 99.9 & 95.9 & 97.6 & 1.3 & 4.7 \\
    & 2 on cloth, 2 on table & 96.6 & 81.1 & 86.9 & 1.7 & 6.7 \\
    & 2 on cloth, 1 on table & 90.9 & 84.8 & 86.7 & 1.5 & 7.7 \\
    & 2 on table & 83.7 & 93.4 & 86.6 & 1.4 & 7.5 \\    
    & 1 on table, 1 on cloth & 90.2 & 84.1 & 85.3 & 1.6 & 7.0 \\
    & 3 on cloth, 1 on table & 90.2 & 77.7 & 81.5 & 2.3 & 5.6 \\
    & 1 on cloth, 1 on table & 82.9 & 77.4 & 77.6 & 1.5 & 8.9 \\
    & 2 on cloth & 82.3 & 75.2 & 76.2 & 1.8 & 7.2 \\
    & 1x folded (2 visible) & 64.2 & 82.6 & 70.5 & 1.3 & 6.1 \\
    & 2x folded (1 visible) & 62.4 & 87.8 & 70.4 & 1.6 & 4.1 \\
    \midrule
    \multirow{10}{*}{\rotatebox[origin=c]{90}{Towels}} & Towel rag & 86.2 & 96.1 & 89.6 & 1.6 & 7.6 \\
    & Checkered rag med. & 92.0 & 90.4 & 89.6 & 1.5 & 7.1 \\
    & Small towel & 94.7 & 85.3 & 87.7 & 1.8 & 8.1 \\
    & Checkered rag big & 92.2 & 85.9 & 86.7 & 2.0 & 5.2 \\
    & Big towel & 85.0 & 91.4 & 86.0 & 1.6 & 7.6 \\
    & Waffle rag & 86.6 & 87.2 & 83.7 & 1.5 & 7.4 \\
    & Linen rag & 78.5 & 88.3 & 80.9 & 1.4 & 6.0 \\
    & Waffle rag stripes & 82.2 & 83.3 & 80.7 & 1.6 & 5.8 \\
    & Cotton napkin* & 80.4 & 84.7 & 80.4 & 1.7 & 5.6 \\
    & Checkered rag small* & 81.5 & 77.8 & 76.7 & 1.5 & 6.3 \\
    \midrule
    \multirow{5}{*}{\rotatebox[origin=c]{90}{Background}} & Festive tablecloth* & 87.1 & 86.4 & 84.5 & 1.6 & 6.7 \\
    & \makecell[l]{Green checkered\\tablecloth} & 85.8 & 85.7 & 83.7 & 1.7 & 5.5 \\
    & White desk & 83.1 & 80.7 & 78.7 & 1.7 & 5.6 \\
    & Poster & 77.7 & 80.1 & 76.4 & 1.5 & 6.5 \\
    & Red tablecloth & 74.8 & 77.0 & 73.1 & 1.6 & 6.7 \\
    \midrule
    \multirow{2}{*}{\rotatebox[origin=c]{90}{Cltr.}} & yes & 82.4 & 83.3 & 80.4 & 1.7 & 6.6 \\
    & no & 84.4 & 85.3 & 82.6 & 1.6 & 6.3 \\
    \bottomrule
  \end{tabular}  
\end{table}

\begin{figure}
    \centering
    \includegraphics[width=0.83\linewidth]{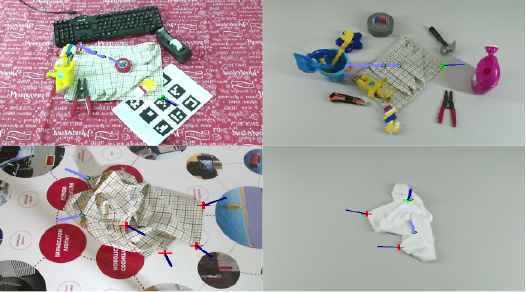}
    \caption{Examples of false and missed grasp point detections (true as a green cross, false as a red cross, and ground-truth as a green dot).}
    \label{fig:detection_false}
\end{figure}

\subsection{Comparison with related methods}

Finally, we conduct a comprehensive benchmark of CeDiRNet-3DoF and several {recent} state-of-the-art models. The evaluation encompasses two categories of approaches {{that are suitable for grasp point detection}: i) point-supervision approaches similar to CeDiRNet and ii) bounding box approaches. {For evaluation, we selected the best-performing recent general methods and two variants of recently proposed key-point detectors tailored for cloth grasping}. In this benchmark, we focus solely on grasp point detection, omitting angle-of-approach accuracy assessment since most of the related approaches do not address it. All results are presented in Table~\ref{tab:related_work}.

All related methods underwent the same training procedure as CeDiRNet-3DoF: pre-training on synthetic images for 10 epochs, followed by at least 10 additional epochs on real images. In certain cases, longer training was necessary for successful convergence. Where applicable, we also consider different backbones for a fair comparison. The methods were exclusively applied to RGB data, as they were specifically designed for this type of data. To align with our method, we apply the same augmentation from CeDiRNet-3DoF to all related methods, utilizing $768\times768$ images for both training and testing.

\begin{table}
  \centering    
  \caption{Related work benchmark on ViCoS {Towel} Dataset. 
  \label{tab:related_work}}
  \setlength{\tabcolsep}{3pt}
  \begin{tabular}{lccccccc}
    \toprule
    \textit{Model} & \textit{Backbone} & \textit{\makecell[c]{Train\\Epoch}} & \textit{Depth} & \textit{\makecell[c]{Prec.\\~[\%]}} & \textit{\makecell[c]{Recall\\~[\%]}} & \textit{\makecell[c]{F1\\~[\%]}} \\
    \midrule
    \midrule
    \multirow{2}{*}{\makecell[l]{CeDiRNet-3DoF\\(our)}} & ConvNext-B & 10 & \checkmark & 83.0 & 84.6 & \textbf{81.4} \\
     & ConvNext-L & 10 & \checkmark & 82.3 & 84.5 & 80.8 \\
    \midrule
    \midrule
    \multirow{2}{*}{\makecell[l]{CeDiRNet-3DoF\\(our)}} & ConvNext-L & 10 & $\times$ & 79.7 & 82.5 & \textbf{78.4} \\
     & ConvNext-B & 10 & $\times$ & 81.5 & 80.2 & 78.0 \\
    \midrule
    \makecell[l]{DINO~\cite{Zhang2023d}} & ConvNext-B & 12 & $\times$ & 72.5 & 79.2 & 72.7 \\
    \midrule
    \multirow{3}{*}{\makecell[l]{DeformDETR\\\cite{Zhu2021}}} & ConvNext-B & 50 & $\times$ & 74.4 & 75.2 & 71.6 \\
     & ResNet101 & 50 & $\times$ & 68.8 & 67.4 & 65.3 \\
     & ResNet50 & 50 & $\times$ & 68.9 & 64.3 & 63.2 \\
    \midrule    
    \multirow{2}{*}{DETR~\cite{Carion2020}} & ConvNext-B & 500 & $\times$ & 65.7 & 62.2 & 61.2 \\
     & ResNet101 & 500 & $\times$ & 64.5 & 59.6 & 59.2 \\
    \midrule    
    \multirow{3}{*}{\makecell[l]{Faster R-CNN\\\cite{He2017}}} & ResNext101 & 10 & $\times$ & 70.0 & 73.1 & 68.3 \\
     & ResNet101 & 10 & $\times$	& 68.6 & 66.7 & 64.4 \\
     & ResNet50 & 10 & $\times$ & 66.6 & 71.3 & 65.8 \\
    \midrule    
    YOLOv7~\cite{Wang2023b} & yolov7x & 20 & $\times$ & 44.8 & 61.0 & 48.3 \\
    \midrule    
    \multirow{3}{*}{\makecell[l]{Cloth grasping\\keypoints\\ \cite{Lips2022,Lips2024}}} & {ConvNext-B} & 50 & $\times$ & 68.6 & 70.6 & 65.7 \\
    & ViT-T~\cite{Lips2024} & 50 & $\times$ & 50.5 & 57.7 & 50.6 \\
    & UNet~\cite{Lips2022} & 50 & $\times$ & 29.3 & 39.5 & 30.7 \\
    \midrule  
    P2PNet~\cite{Song2021} & VGG16 & 20 & $\times$ & 46.7 & 53.8 & 46.2 \\
    \midrule    
    \multirow{4}{*}{\makecell[l]{CenterNet++\\\cite{Duan2022a}}} & ResNext101 & 24 & $\times$ & 50.6 & 56.2 & 49.4 \\
     & ResNet101 & 24 & $\times$ & 48.0 & 45.5 & 43.0 \\
     & ResNet50 & 36 & $\times$ & 39.5 & 42.4 & 37.0 \\
     & Swin-L & 24 & $\times$ & 43.9 & 33.1 & 35.1 \\
    \midrule
    \multirow{3}{*}{\makecell[l]{CenterNet\\\cite{Zhou2019}}} & ResNet101 & 150 & $\times$ & 30.7 & 37.7 & 30.6 \\
     & ResNet101 & 50 & $\times$ & 26.6 & 32.8 & 26.4 \\
     & Hourglass-101 & 50 & $\times$ & 25.2 & 30.8 & 24.9 \\
    \midrule    
    PET~\cite{Liu2023}  & VGG16 & 20 & $\times$ & 32.0 & 19.1 & 22.2 \\
    \bottomrule
  \end{tabular}  
\end{table}

\paragraph{Point-supervision approaches} We assessed the performance of CenterNet~\cite{Zhou2019}, CenterNet++~\cite{Duan2022a}, PET~\cite{Liu2023}, and P2PNet~\cite{Song2021} as general methods, 
 {and two variants of a specialized method designed for grasp point detection on cloths using key-points~\cite{Lips2022,Lips2024}}. The results are detailed in the lower rows of Table~\ref{tab:related_work}. CenterNet and CenterNet++, akin to CeDiRNet-3DoF, focus on regressing the center positions of objects. Their capacity to regress bounding boxes was not utilized in our benchmark. Both related methods exhibit significantly inferior performance compared to CeDiRNet-3DoF, with 30\% and 49\% in F1 measure for CenterNet and CenterNet++, respectively, while CeDiRNet-3DoF surpasses them with 78\%. The change in backbones to Swin-L or Hourglass-101 did not yield improvements. 

PET and P2PNet, designed for point localization in crowd counting applications, demonstrate similarly subpar results, {with 22\% and 46\% F1 measure, respectively, for PET and P2PNet}, respectively, markedly worse than CeDiRNet-3DoF. We attribute the poor results to the combination of the backbones (VGG16) and the methods' design for crowd counting, where the loss function anticipates a large number of object locations. In our case, the detection focuses on a maximum of four points in each image, creating a significantly unbalanced set of points compared to the number of background pixels.

{ Purpose-build key-points methods for grasp point detection on cloths~\cite{Lips2022,Lips2024} also demonstrated subpar performance. A variant from ICRA 2022 workshop~\cite{Lips2022} with U-Net achieved only 30\% in F1 measure, while the latest RA-L 2024 journal variant~\cite{Lips2024} achieved 50\%. However, the best performance was achieved with further changes to the backbone, attaining 66\% F1 measure when using ConvNext-B. Nevertheless, this still lags behind CeDiRNet-3DoF by 12 pp.}

\paragraph{Bounding box approaches} As bounding box methods we included Faster R-CNN~\cite{He2017}, YOLO~v7~\cite{Wang2023b}, DETR~\cite{Carion2020}, DeformDETR~\cite{Zhu2021}, and DINO~\cite{Zhang2023d}, with the latter three representing state-of-the-art transformer approaches. As a training bounding box, we used a fixed-sized rectangle ($50\times50$ pixels) centered over the annotation point. For testing, the center of the predicted bounding box was utilized as the final point prediction.

The least-performing bounding box method was YOLO v7 (48\% F1), comparable to CenterNet++ but significantly lagging behind CeDiRNet-3DoF (78\% F1). Faster R-CNN with a ResNext101 backbone yielded much better results (68\% F1). Among transformer-based methods using ConvNext-B, DETR achieved a maximum F1 measure of 61\%, while DeformDETR and DINO demonstrated slightly superior performance compared to Faster R-CNN, with F1 measures of 71\% and 72\%, respectively. Notably, CeDiRNet-3DoF, employing the same ConvNext-B backbone, outperformed all related methods, surpassing the best-performing related works, DINO and DeformDETR, by 6-7 pp. Furthermore, this superior performance was achieved with only 10 epochs, whereas DeformDETR required 50 epochs, and DETR needed 500 epochs to converge. Notably, the performance improves even further when depth data is added to CeDiRNet-3DoF.

\section{Conclusion} \label{sec:conclusion}

In this work, we introduced a novel approach for grasp point detection on deformable cloth objects using CeDiRNet-3DoF. Leveraging a re-parameterization of center directions and a dedicated localization module, the proposed method demonstrated robust performance across diverse towel types, configurations, backgrounds, and clutter scenarios. A significant contribution of this work is the introduction of the publicly available ViCoS {Towel} Dataset, comprising 8,000 real-world and 12,000 synthetic images, which serves as a valuable resource for training and evaluating data-dependent deep-learning approaches. 

The proposed method has been extensively evaluated on the ViCoS {Towel} Dataset. Through an ablation study, we highlighted {the benefits of our design choices} and the significance of synthetic data and depth information. Our parametric performance study provided further nuanced insights into performance across different towel types, corner configurations, backgrounds, and clutter scenarios. Notably, the method excelled in scenarios where cloth is laid flat on the table with all four corners well-visible, achieving over 97\% F1. In a comprehensive benchmark against several related methods, CeDiRNet-3DoF outperformed both point-supervision and bounding box approaches, including the latest transformer-based models, DINO and DeformDETR, {and a purpose-build key-point detector for cloth}, highlighting its effectiveness and efficiency.

Our findings underscore the potential of CeDiRNet-3DoF for real-world robotic manipulation tasks. Despite achieving excellent results on towels with well-visible corners, this work highlights challenges for folded and crumpled objects where results are still not perfect. {Method may also face limitations due to the need for high output resolution when grasp points are close together, potentially making them hard to distinguish in sin/cos fields. Another constraint is its focus on point detection, with the dataset featuring grasp points exclusively at physical corners, neglecting edges useful for cloth manipulation. These challenges present opportunities for future research. We aim to expand the dataset by annotating towel edges in upcoming studies, thereby providing further data for advancing learning-based methods for grasping point localization on cloth.

\section*{Acknowledgments}
Acknowledgments: This work was in part supported by the ARIS research projects J2-3169 (MV4.0) and J2-4457 (RTFM) as well as by research programme P2-0214.

{
\bibliographystyle{IEEEtran}
\bibliography{library}
}

\end{document}